\documentclass[9pt,twocolumn,twoside,lineno]{pnas-new}
\templatetype{pnasresearcharticle} 

\usepackage{EvolvingAI}
\newif\ifcomments

 \commentstrue

\ifcomments
\newcommand{\comments}[1]{#1}
\else
\newcommand{\comments}[1]{}
\fi

\title{Continual learning under domain transfer with sparse synaptic bursting}

\author[a,d]{Shawn L. Beaulieu}
\author[b,e,f,g]{Jeff Clune}
\author[c,d]{Nick Cheney} 

\affil[a]{shawn.beaulieu@uvm.edu}
\affil[b]{jclune@gmail.com}
\affil[c]{ncheney@uvm.edu}

\affil[d]{Department of Computer Science, University of Vermont: Burlington, VT, USA}
\affil[e]{Department of Computer Science, University of British Columbia: Vancouver, BC, Canada}
\affil[f]{Vector Institute}
\affil[g]{Canada CIFAR AI Chair}

\leadauthor{Beaulieu} 

\begin{abstract}

AI programs with the intelligence, resilience, and autonomy approaching that of biological systems must be capable of learning and retaining new information without arbitrarily frequent re-training. In this paper, we introduce a system that can learn sequentially over previously unseen datasets (ImageNet, CIFAR-100) with little forgetting over time. This is done by controlling the activity of weights in a convolutional neural network in a context-dependent manner using top-down regulation generated by a second feed-forward neural network. We find that our method learns continually under domain transfer to a new dataset with sparse bursts of heavy-tailed activity in weights that are recycled across tasks, rather than by maintaining task-specific modules. Sparse synaptic bursting is found to balance activity and suppression such that new functions can be learned without corrupting extant knowledge, perhaps mirroring the balance of order and disorder in systems poised at the edge of chaos. This behavior emerges during a prior pre-training (or ``meta-learning'') phase in which regulated synapses are selectively disinhibited, or grown, from an initial state of uniform suppression through prediction error minimization.  

\end{abstract}

\begin{document}
\maketitle
\thispagestyle{firststyle}
\ifthenelse{\boolean{shortarticle}}{\ifthenelse{\boolean{singlecolumn}}{\abscontentformatted}{\abscontent}}{}


\dropcap{C}atastrophic forgetting is the phenomenon wherein an artificial neural network trained over a sequence of inputs loses its ability to perform a function it acquired earlier in the sequence as new information is learned \cite{french1999, Kirkpatrick2017}. This happens because the ability of a network to perform a given function is determined by its current state, which includes the arrangement and values of its synapses. Changes of state, as a consequence of learning, may corrupt functions that are not robust to such perturbation. Forgetting is typically overcome by scrambling the temporal structure of learning with large batches of randomly ordered inputs that are stored at the programmer’s discretion (IID training). This eliminates the problem of forgetting because no class of input is restricted to a single segment of the training sequence. All inputs are equally likely to be seen by the network at any point in time, and so cannot be forgotten as such. But it is unrealistic for most real-world applications to assume that all relevant data can be obtained prior to model deployment. Additionally, we may encounter constraints on time, storage, and compute that prevent us from scaling traditional solutions to catastrophic forgetting ~\cite{Kirkpatrick2017, kudithipudi2022biological, hayes2019memory, hayes2021replay}. 

In this paper we present an algorithm for continual learning in a convolutional classifier, whose synapses are regulated in a context-dependent manner by a second neural network, which itself undergoes continual change. Prior to learning continually, synaptic regulation is first pre-trained via meta-learning~\cite{Finn2017, Javed2019}. We find that meta-learned regulation which best avoids catastrophic forgetting is initialized during pre-training to be uniformly \textit{suppressant}. For the model to acquire useful functions, the regulator must learn to \textit{disinhibit}, or "grow", synapses whose activation helps to identify relevant features of the input. After meta-learning to grow sensors in the classifier, both the regulator and classifier are greedily updated over long sequences of input sampled from a previously unseen dataset (\textit{domain transfer}). Under domain transfer from Omniglot \cite{Lake2015} to ImageNet \cite{Russakovsky2015} and CIFAR-100 \cite{Krizhevsky2009LearningML}, we find that the regulator avoids catastrophic forgetting by inducing sparse bursts of heavy-tailed activity in synapses that are recycled across tasks, rather than by maintaining task-specific modules \cite{Kirkpatrick2017, Javed2019, Beaulieu2020, Ellefsen2015, Masse2018}. We propose that sparse synaptic bursting allows the regulator to control the \textit{amount} of activity (and, thus, plasticity) in the classifier, rather than its task-modular \textit{location}, such that prior knowledge is protected without blocking adaptation to new inputs. 

\begin{figure}[!t]
\includegraphics[width=1\linewidth]{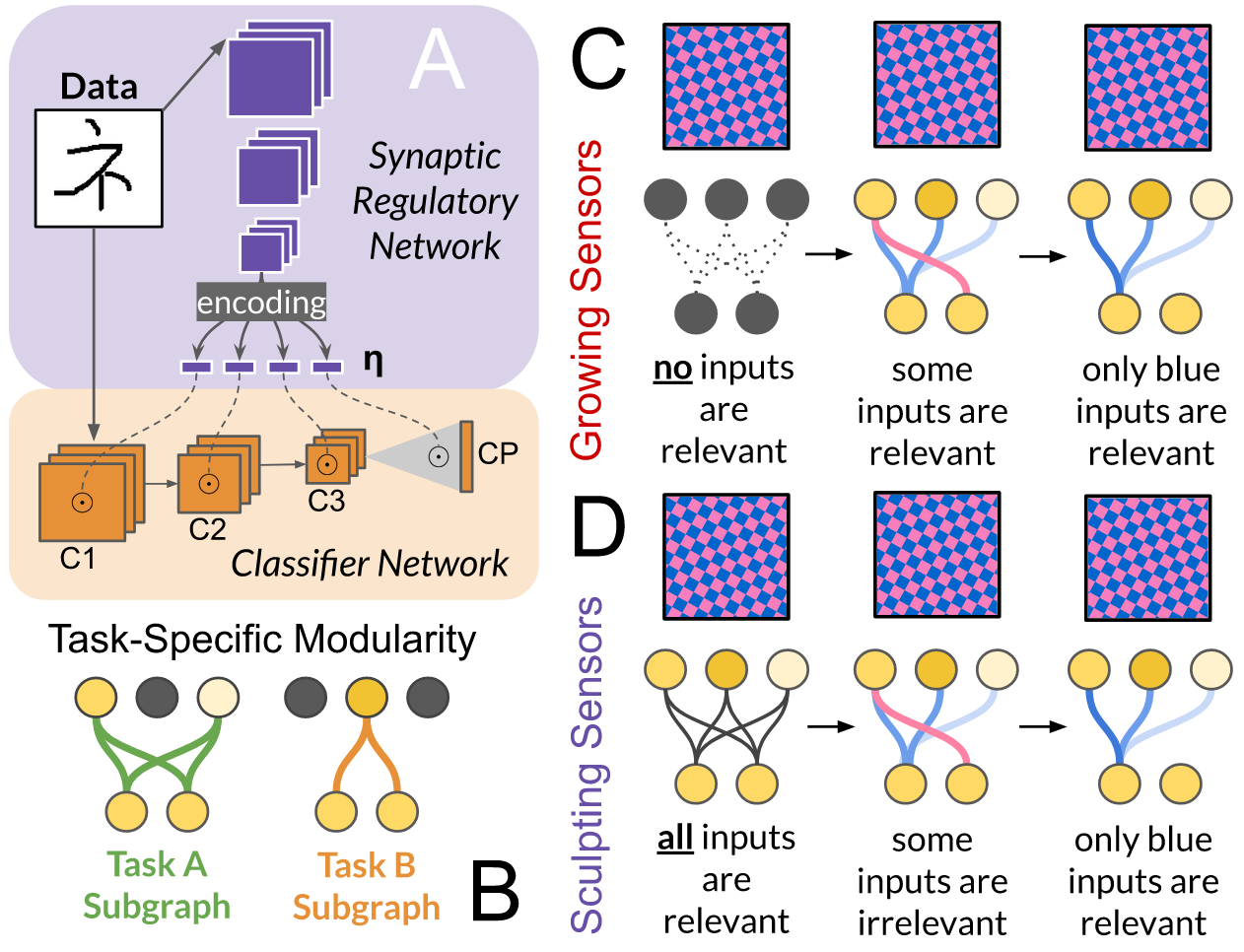}
\caption{(\textbf{A}) Architecture for Tuning Synapses via Allostatic Regulation ("TSAR", Materials and Methods). (\textbf{B}) Task-specific modularity has traditionally be used to overcome catastrophic forgetting. Disjoint subgraphs, or "modules",, prevent prediction interference and unwanted weight changes (\textbf{C}) By initializing meta-learned regulation to be highly suppressant (``Grow'', Materials and Methods) we establish a prior on regulation such that no input features are relevant to the recruitment of synapses in the classifier. For performant behavior to emerge, the regulator must learn to disinhibit or ``grow'' synapses which minimize prediction error. Conversely, (\textbf{D}) if regulation is initialized to be highly permissive (``Sculpt'', Materials and Methods) then all input features initially are specified as relevant.}
\label{fig:concept}
\end{figure}

\begin{figure*}
\centering
\includegraphics[width=1\linewidth]{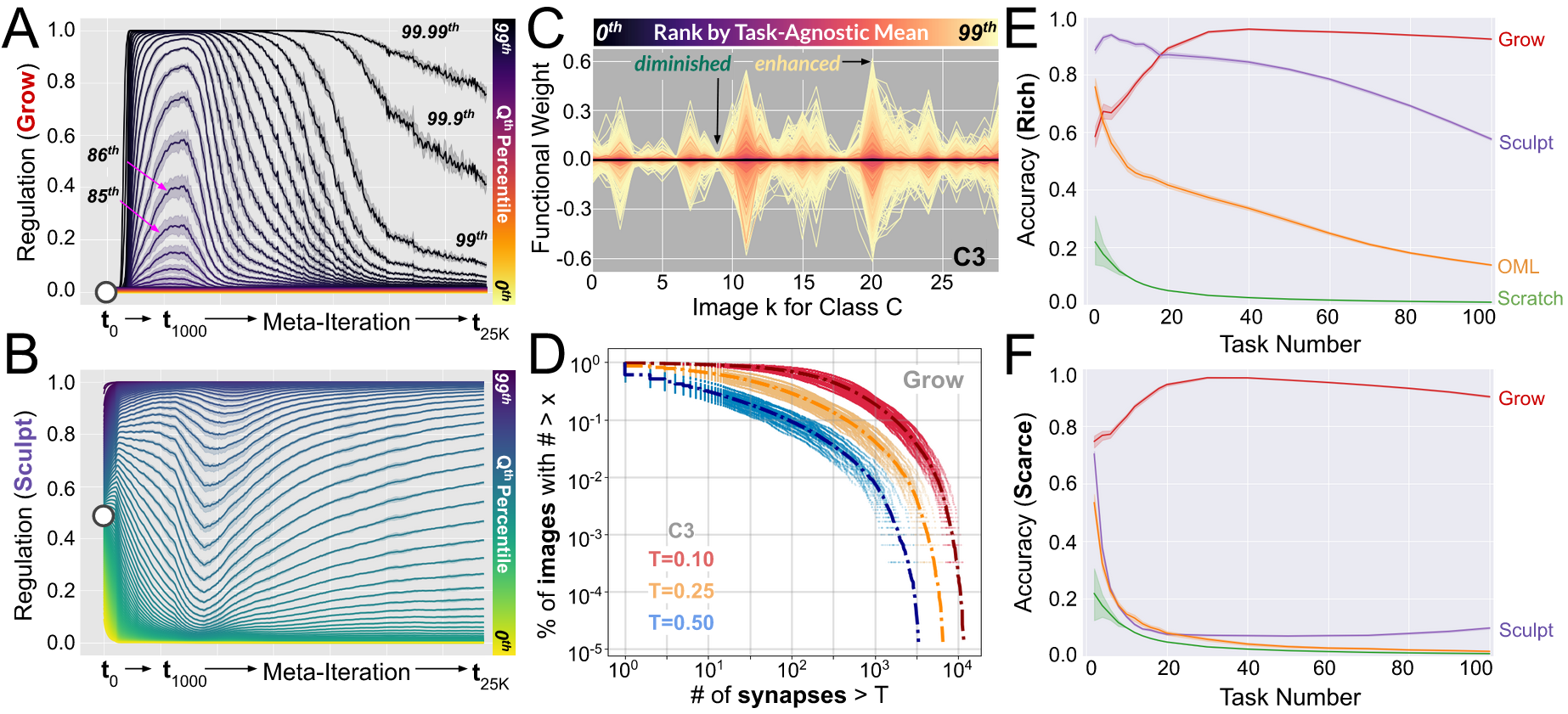}
\caption[width=1\linewidth]{Regulation of layer C3 during meta-learning of Grow (\textbf{A}) and Prune (\textbf{B}). We report the $Q^{th}$ percentile of regulation across 250 randomly sampled meta-learning classes. Confidence intervals are computed across runs (lower bound=$20^{th}$ percentile; upper bound=$80^{th}$ percentile). (\textbf{C}) Randomly sampled window of synaptic activity (post-masking) under domain transfer for Grow. (\textbf{D}) Complementary cumulative distribution function (CCDF) for the percent of images in ImageNet that cause a given number of synapses to receive regulation above a threshold (T). We report individual runs (dot) and the mean across runs (dash-dot). Continual learning under domain transfer to Imagenet after meta-learning on 100\% (\textbf{A}) or less than 3\% of Omniglot classes (\textbf{B}). The Grow treatment (initial regulatory bias=-8)
outperforms the Sculpt treatment (bias=0) 
on the domain transfer task (\textbf{C}) and this difference is exacerbated with data limited meta-learning.}
\label{fig:performance}
\centering
\end{figure*}

Our analysis suggests that an apposite balance of activity and suppression, in which synaptic activity is permitted with inverse proportion to its magnitude, is achieved only by those regulators set to a specific range of initial pre-training conditions: when meta-learned regulation is initialized to be more suppressant or permissive relative to the optimal range, performance under domain transfer falls—often catastrophically. Sparse synaptic bursting is found to exist in meta-learned models forced to disinhibit initially suppressed weights (sensor \textit{growth}), and is absent in meta-learned models forced to suppress initially disinhibited weights (sensor \textit{sculpting}). We observe that meta-learned sensor growth is characterized by a sharp disinhibition of initially suppressed synapses, followed by a protracted phase of pruning—culminating in a heavy-tailed distribution of synaptic recruitment. This mirrors what has been observed for neural development in biological systems, for which initially sparse connectivity gives way to increased synaptic density in early adolescence, and subsequent pruning of excess connections during teenage and adult years \cite{Neniskyte2017, Millan2018}. We show that meta-learned sensor growth learns without forgetting under domain transfer even when the amount of meta-learning data is reduced to less than 3\% used in the original treatment.

\section{Tuning synapses via allostatic regulation.}\label{sec:Section 1}

The method we introduce consists of two neural networks: (\textit{i}) a convolutional classifier network, consisting of three convolutional layers and a linear class prediction layer; and (\textit{ii}) a feed-forward regulatory network that takes in the same input image as the classifier network, but generates a real-valued mask on every weight of the classifier (sigmoidal output in the range [0, 1]). This mask affects both forward propogation of inputs and the backpropagation of error, enabling context-dependent selective plasticity ~\cite{Masse2018, Beaulieu2020}. Thus, our model has kinship with fast weight memory systems \cite{Ashby1960, Schmidhuber1992, Ba2016, Tsuda2020} for which information is dynamically routed through a quickly evolving network by a supervising controller that adapts more slowly. However, we do not explicitly encode operations for maintaining a short-term memory, except insofar as previously learned weights are preserved and recruited by the regulator. Our model therefore belongs to the class of continual learning systems that learn how to learn according to empirical success \cite{Stanley2002, SynapticIntelligence, Finn2017, NeuralArchSearch} rather than manually designed rules \cite{IseleCosgun2018, Fernando2017PathNetEC, French1992, Ellefsen2015, Kirkpatrick2017, InductiveBias, Zhang2017}. 

One of the chief contributions of this work is to demonstrate the relative advantage of systems that are forced to grow over time \cite{Kauffman1986, Watts1998, Sporns2004, Bongard2011, Bernatskiy2015, Neniskyte2017, GrowingNeuralNets} by meta-learning how to activate uniformly suppressed weights, against those that are forced to sculpt away components ~\cite{LeCunoptimalbrain, SynFlow, LotteryTicket, Bengio2013EstimatingOP, zhou2019deconstructing, wortsman2020supermasks} by meta-learning how to suppress uniformly active weights (Section IV).

\section{Method for meta-learned sensor growth.}\label{sec:Section 2}

In selecting for the ability to learn without forgetting, we first pre-train TSAR (Tuning Synapses via Allostatic Regulation) using the Online-aware Meta-Learning (OML) algorithm \cite{Javed2019}. This involves an inner loop of sequential learning over training images sampled from a single Omniglot class~\cite{Lake2015}, followed by an outer loop update on a batch of images sampled from $K$ other random classes. The inner loop updates the classifier only, while the outer loop updates the initial weights of both the regulator and classifier, which are then used to start the next inner-loop. As a result, our model must learn how to learn over a sequence of Omniglot images, such that the corresponding updates do not corrupt the ability to classify previously seen Omniglot images. In addition to the baseline setting that uses 963 Omniglot classes (``data rich''), we also present a ``data scarce'' setting containing less than 3\% of this data, or just 25 randomly sampled Omniglot classes. 

Finally, we obtain regulators that ``grow'' connections in the classifier by varying the strength of the initial bias on regulation prior to meta-learning. In this way, we can cause regulation to be more or less \textit{permissive} ($\approx 1$, unmasked) or \textit{suppressant} ($\approx 0$, fully masked). We perform a sweep over initial biases covering the range of integers from $[-12,12]$, and find that optimal performance is obtained with a bias of -8 (Fig.~\ref{fig:performance}, C). This initialization we refer to as the canonical ``Grow'' condition, which we set against the canonical ``Sculpt'' condition, having the standard initial bias of 0. Sculpt begins meta-learning with weights in the classifier being uniformly active, and which may or may not be sculpted or pruned away over the course of meta-learning. For more detailed information, see Materials and Methods.

\begin{figure*}
\centering
\includegraphics[width=1\linewidth]{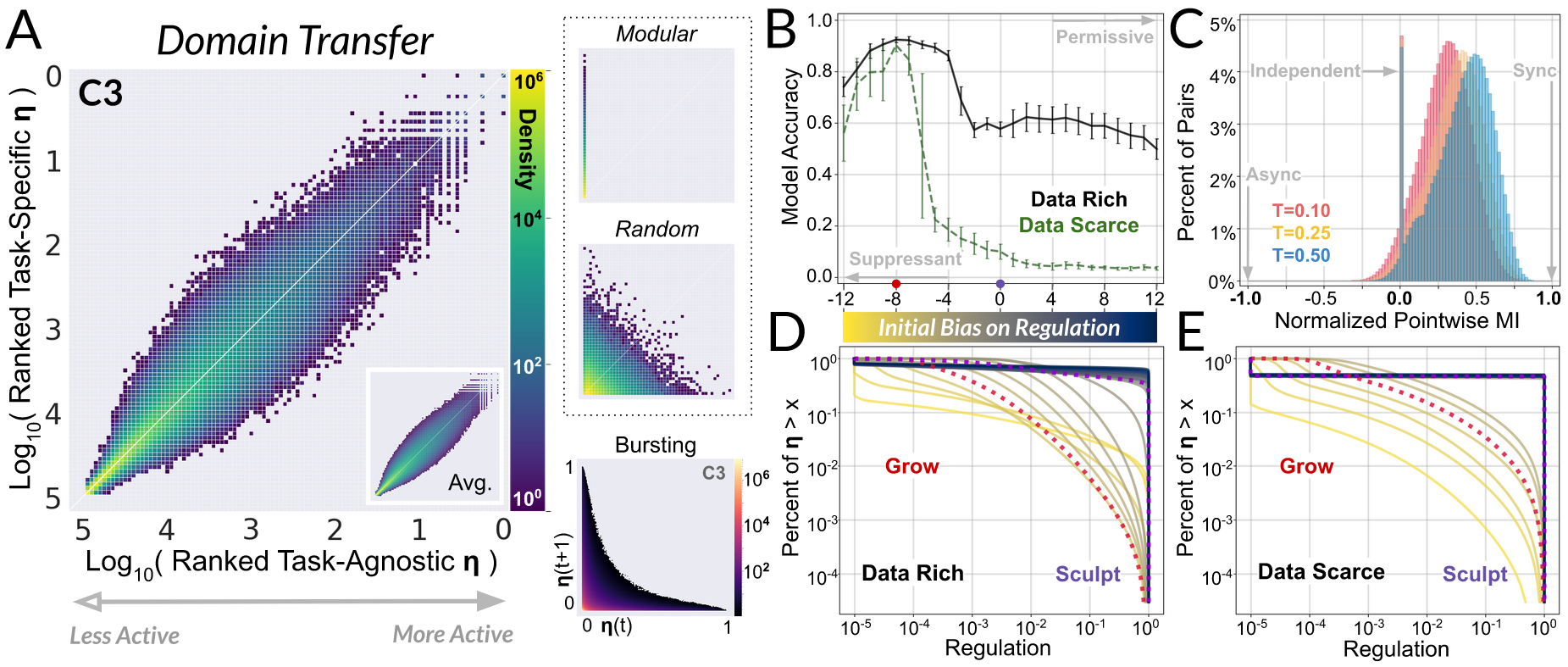}
\caption[width=1\linewidth]{
Regulation of synapses under domain transfer does not show evidence of task-specific modularity (see hypothetical ``Modular'') but is largely task-agnostic (\textbf{A}). Synapses are likely to be suppressed even if they were highly active on the previous image from the same class (``Bursting''). (\textbf{B}) Performance under domain transfer to ImageNet for the range of initial biases considered. Standard deviation is reported for error bars. (\textbf{C}) Normalized pointwise mutual information \cite{churchhanks} for each pair of synapses receiving regulation above the given threshold for more than 1\% of images. We find a positive bias in the degree of correlation in synapse pairs, but an absence of strong synchronicity or asynchronicity. This indicates that the coalition of weights used for a given image is relatively heterogeneous with respect to those used for other images. Increasingly scale-free cumulative distribution functions for regulation obtain under domain transfer for increasingly successful regulators after data-scarce (\textbf{E}) and, especially, data-rich (\textbf{D}) meta-learning. See Appendix 1 for domain transfer to CIFAR-100. 
}
\label{fig:regulation}
\centering
\end{figure*}

\section{Performance under domain transfer.}\label{sec:Section 3}

Prior studies test the ability of meta-learned neural networks to avoid catastrophic forgetting on previously unseen data taken from the same domain used for meta-learning, thereby limiting the extent to which anything new must be learned \cite{Javed2019, Beaulieu2020}. Instead, we subject our meta-learned model to input sequences from entirely new \textit{datasets}. The model meta-learns on the Omniglot image dataset \cite{Lake2015}, and is tested for catastrophic forgetting on input sequences taken from ImageNet \cite{Russakovsky2015} and CIFAR-100 \cite{Krizhevsky2009LearningML}. Doing this is analogous to the transfer of robots from simulation to reality \cite{Mouret2017}, or the introduction of a novel stressor in biology \cite{EmmonsBell2019}. Under domain transfer, we train on a random sequence of 100 previously unseen tasks, where each task contains 30 randomly sampled images from a pool of 600 possible images for that class. Each run then consists in a total of 3,000 sequential gradient updates. Accuracy is measured as the degree to which prior inputs can be recalled. This captures the phenomenon of forgetting by quantifying how much of what was actually learned is remembered over time \cite{Javed2019, Schmidhuber1992}(Fig.~\ref{fig:performance} A,B). We find that Grow (92.2$\% \pm$1.1$\%$) learns sequentially under domain transfer without significant loss in performance as new tasks are encountered—while competing methods, including different regulatory initializations, forget catastrophically (Sculpt: 57.5$\%\pm$2.8$\%$). To calibrate our understanding of what it means to perform well on this problem, we compare against the meta-learned algorithm \textit{OML} \cite{Javed2019}, as well as our classification network trained from scratch without meta-learning or regulation (\textit{Scratch}) both of which perform poorly, indicating the problem is non-trivial (OML: 13.87$\% \pm$0.729$\%$, Scratch: 0.971$\% \pm$0.10$\%$). For the data scarce condition, performance of Grow modestly declines relative to the data rich condition, while Sculpt, OML, and Scratch forget catastrophically (Figure 1B). Similar results obtain for domain transfer to CIFAR-100 with minor variations in accuracy (Appendix 3A).

\section{Task-specific modularity.}\label{sec:Section 4}

All analysis henceforth concerns regulation of the third convolutional layer of the classification network (C3) unless otherwise stated. Qualitatively similar results were obtained for the regulation of other convolutional layers (Appendix 5-13).

Past efforts to solve catastrophic forgetting have relied on task-specific modules for conditionally modifying disjoint sets of weights \cite{Ellefsen2015, Kirkpatrick2017, Javed2019, Beaulieu2020}. To determine whether regulation in TSAR is calibrated for such task-specific modularity (Figure 1B), we reason that task-specific modules should be characterized by heightened synaptic activity within a given task but reduced activity for all other tasks. Task-Specific activity for synapse $i$ in layer $\ell$ is computed as the average regulatory signal governing that synapse, $\eta_{i}^{\ell}$, over all $K$ instances of a given task, $C$ (Eq.\ref{eq:2}). Task-Agnostic activity is computed as the average Task-Specific activity over all tasks not equal to $C$ (Eq.\ref{eq:3}). In accordance with prior work \cite{LeCunoptimalbrain, SynFlow}, we take activity to be a proxy for weight salience, and therefore regard ranked activity as ordered importance. For Grow, regulation is heavy-tailed (Figure 3D); as such, we are concerned with \textit{logarithmic} rank \cite{Dodds2020}. Task-specific modularity, or its absence, can then be visualized by plotting ranked Task-Specific activity against ranked Task-Agnostic activity for each task under domain transfer (Figure 3A).

\begin{equation}\label{eq:2}
    \operatorname{Task-Specific \ Activity} = \frac{1}{K} \sum_{k=0}^{K} \eta_{i}^{\ell}(C_{k})
\end{equation}

\begin{equation}\label{eq:3}
    \operatorname{Task-Agnostic \ Activity} = \frac{1}{T-1} \sum_{t \neq C} \ \frac{1}{K}{\sum_{k=0}^{K} \eta_{i}^{\ell}(t_{k})}
\end{equation}

We find that the synapses which are the most active on a given task are among the most active overall. Indeed, the majority of weights rank similarly over all tasks as they do for any individual task. Finally, we do not find that any synapse which ranks high for any given task also ranks low for all other tasks. Thus, we do not see evidence for task-specific \textit{modularity} in the output of the regulatory network.

\begin{figure*}
\centering
\includegraphics[width=1\linewidth]{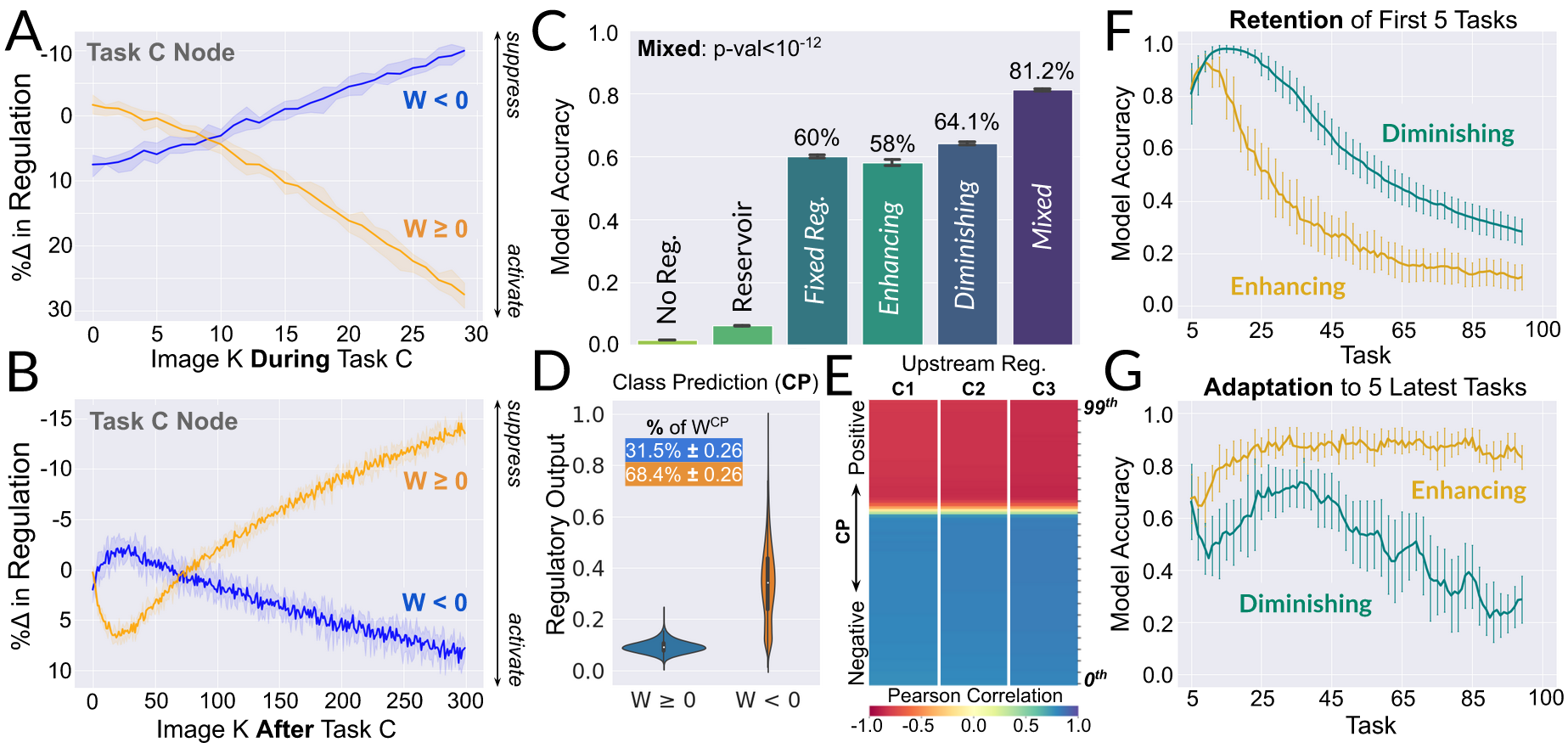}
\caption[width=1\linewidth]{
Within each task, positive weights leading into the correct output node become increasingly active and thus increasingly plastic (\textbf{A}), and then transition over the next 10 tasks to become increasingly dormant and thus increasingly protected from future weight changes that may lead to forgetting (\textbf{B}). (\textbf{C}) For each class in ImageNet, ``enhancing'' images are defined to be the top-30 images for that class ranked by mean regulatory output, while ``diminishing'' images are defined to be the bottom-30 images for that class ranked by mean regulatory output. Training over a sequence of 100 classes populated exclusively by ``enhancing'' or ``diminishing'' images results in significantly lower performance than training over 100 consecutive classes with randomly ordered but even mixtures of enhancing and diminishing images. (\textbf{D}) In the class prediction layer (CP) positive weights receive regulation that is more suppressant and lower variance, while negative weights receive regulation that is more permissive and high variance. (\textbf{E}) Positive weights in CP are negatively correlated with upstream regulation, such that when upstream spiking occurs, positive weights are suppressed. Conversely, negative weights in CP are positively correlated with upstream regulation. Pearson correlation is computed with respect to first differences for (i) \textit{upstream activity}, defined as the mean regulation for the top 10\% of weights ranked by regulation; and (ii) the mean regulation governing weights in CP ranked by weight value and grouped centile. The top row of then reports the Pearson correlation between the mean regulation going to the top-10\% of weights in C1, C2, or C3 and the mean regulation going to the 99$^{th}$ percentile of weights in CP. We plot the mean Pearson correlation per centile across all 25 models. Diminished processing is found to improve performance \textit{retention} (\textbf{F}) while enhanced processing is found to facilitate adaptation to new inputs (\textbf{G}). Error bars report standard deviation across all 25 models. 
}
\label{fig:cp_regulation}
\centering
\end{figure*}

\section{Synaptic bursting.}\label{sec:Section 5}

In the absence of task-specific modularity, we observe that synapses are conditionally recruited for task-agnostic bursting, characterized by transient activity in response to particular \textit{images} rather than particular \textit{tasks}. This produces waves of \textit{enhanced} and \textit{diminished} processing of inputs, in which some images strongly use and modify many weights of the network, while other images weakly use and modify very little of the network (Figure 2C). Yet strongly correlated activity in pairs of synapses across all images is rare, indicating variable spike composition (Figure 3C). Such weak synchronization may reflect the creation and storage of robustly distributed memories \cite{Wagner2005}, as no single weight, or set of weights, encodes a unique function \cite{Anderson2014, Bruineberg2019, Maass2004, Battaglia2017, Stringer2019} but instead performs multiple functions depending on the context in which it is active. Analysis shows that increasingly performant models exhibit regulation under domain transfer that is increasingly \textit{scale-free} ~\cite{Stumpf2012, Kauffman1991, Kauffman2004, Shew2011, Bertschinger2004, Ramo2007, Sporns2004, Mora2011}. Here scale-free regulation means that inputs which use (unmask) an exponentially larger amount of the classifier are exponentially more rare. We find that images which activate a given \textit{number} of synapses are similarly distributed (Figure 2D). We therefore claim that performant regulation is calibrated to control the \textit{amount} of sensory processing in the classifier, rather than the task-modular \textit{location} where such processing occurs. This is accomplished by maintaining an apposite balance of activity and suppression—and, thus, an apposite balance of plasticity and stability. Comparatively poor regulators, associated with more permissive initial regulation (i.e. Sculpt), do not exhibit this behavior. 

\section{Regulating class predictions.}\label{sec:Section 6}

If upstream sensory processing is characterized by task-agnostic synaptic bursting, how is it that correct class predictions are acquired and maintained by the network? First, we observe that regulation of the class prediction layer (CP) produces markedly different distributions for positive and negative weights (Figure 4D). Positive weights in CP are held at relatively low values with infrequent variation, while negative weights in CP exhibit comparatively high variation around a larger mean value. At the same time, positive and negative weights in CP were found to differ in their correlation with upstream activity, such that negative weights are strongly synchronized with upstream bursting, but positive weights are \textit{inversely} synchronized (Figure 4E). This manner of coordination across layers suggests that different modes of behavior are associated with enhanced and diminished states, although it is unclear precisely what they entail. Future work may shed light on this phenomenon by, for example, studying its relation to rhythmic sampling in biological systems \cite{Golding2002, Fiebelkorn2013, Fiebelkorn2019}. 

We then tracked the average change in regulation governing the positive and negative weights of a given task node, C, over two non-overlapping windows in time: (\textit{i}) when task C is being learned; and (\textit{ii}) when the next 10 tasks that follow task C are learned (Figure 4A,B). This analysis covers the first 90 classes seen under domain transfer for all runs. We find that when task C is learned, the positive weights innervating task node C are gradually \textit{activated}, while the negative weights innervating task node C are gradually \textit{suppressed} (Figure 4A). But when future tasks are encountered this trend is inverted: positive weights are suppressed while negative weights are activated (Figure 4B). This is consistent with the encoding of task-specific information in the positive weights of the relevant task node. Task-specific information is then protected under task-switching with the relative suppression of positive weights in C and the relative activation of negative weights in C. This both preserves the information encoded by the positive weights of task node C, and dampens the excitation of node C when new tasks are encountered.

Together these results indicate how task-specific predictions are acquired and maintained alongside task-agnostic upstream bursting. However, we find that optimizing the regulation of CP under domain transfer is not, on its own, sufficient to achieve good performance: when the weights producing regulatory output for convolutional layers (C1,C2,C3) are fixed, but regulation of CP is trainable, performance drops relative to the condition where regulatory output for all layers is trainable (Figure 4C, ``Fixed''). Nor does meta-learning the classification network without regulation result in effective domain transfer (Figure 4C, "No Reg.", 1.5$\% \pm$0.20$\%$). From this we conclude that optimizing weights and regulation for \textit{all layers} is necessary for successful domain transfer, despite increased susceptibility to catastrophic forgetting. 

\section{Conclusion and Discussion.}\label{sec:Section 7}

In the preceding sections we presented an algorithm for regulating the synapses of a convolutional neural network for continual learning over sequences of previously unseen datasets. This occurs by meta-learning to disinhibit synapses initialized to a state of uniform suppression.

Under domain transfer, we found that regulation does not elicit task-specific modularity, but instead induces sparse bursts of activity in weights that are recycled across tasks. Sparse synaptic bursting is found to be heavy-tailed in both the total distribution of regulatory output (Figure 2B,C) and in the number of synapses it causes to burst image-to-image (Figure 1D). Balancing contrasting states, like activity/suppression and order/disorder, has previously been implicated in the discovery of optimal communication and memory in simulated networks \cite{Kauffman2004, Shew2011, Bertschinger2004, Ramo2007}, and is believed to play a role in the adaptive behavior of living systems \cite{Sporns2004, Mora2011}. While prior work has demonstrated the various benefits of burst-dependent plasticity~\cite{lisman1997bursts, hahn2019portraits, park2019fast, payeur2021burst} bursting itself it is not yet an established component in theories of how to avoid catastrophic forgetting. However, we note a possible connection to work in biological systems on rhythmic sampling \cite{Fiebelkorn2013, Fiebelkorn2019} and distributed robustness \cite{Wagner2005, Anderson2014, Bruineberg2019, Maass2004, Battaglia2017, Stringer2019}. We hypothesize that the initial biases on meta-learned regulation that achieve domain transfer corresponds to a critical range in which regulation learns to balance activity and suppression, thereby enabling adaptation to new inputs without corrupting extant functions. This resolves the core dilemma of continual learning: too much change causes forgetting, while too little change induces brittleness.

We might also analogize regulation of the \textit{amount} of sensory processing, rather than its task-modular \textit{location}, to the phenomenon of 'virtual governors' \cite{wieneCybernetics, Pezzulo2016} for which emergent relational properties, like temperature and pressure, offer more causally effective targets for intervention than the elements that compose them \cite{Hoel2013}. A balance of coarse-grained regulatory states that are insensitive to the precise details of context, like enhanced and diminished perception, could secure greater regulatory efficacy than more fine-grained methods of control. At the same time, component \textit{recycling} is one of the signatures of living systems \cite{Nicholson2019}, and has been proposed to distinguish perception as a mediator of \textit{interaction} \cite{Sporns1994, Anderson2014, Bruineberg2019} rather than a system for constructing accurate world models \cite{ha2018worldmodels}.

Among the factors responsible for the resilience of living systems is the ability to impose meaning on the world through the negotiation of bodily form and function \cite{Pask1962, Cariani2007, Kirchhoff2018, informationSelfStructuring, Pharoah2020,Friston2012, Bruineberg2014}. Unlike machine learning programs, living systems act and perceive in ways that cannot be understood as transmitting information over anatomically fixed, yet tunable, channels \cite{Logan2012, Bongard2021, Nicholson2019}. Rather, the means by which information is made and transmitted are themselves subject to purposive modification. Thus, encoded in the very structure of living systems is that which counts as meaningful \cite{Pfeifer2007, Man2019, merleau-ponty1983, bodybookBongard, Kirchhoff2018}. For neural networks, intelligence is simply what it means to usefully map a given set of inputs to a given set of outputs, but the space of possible actions and percepts typically isn't subject to change—and if it is, the manner in which it can change is rigidly specified in advance \cite{Cariani2007, Bongard2011, Bernatskiy2015, GrowingNeuralNets, Jaafra2019, Stanley2002, Schmidhuber1992, Ba2016, greff2020binding, schlag2021linear}. However, autonomous solutions to problems like catastrophic forgetting require that programs decide according to their own internal logic, which must itself be amenable to change, what information is relevant to the tasks they encounter \cite{Bell1995, Fukushima1988, Fry2002, Javed2019, Beaulieu2020}. This cannot be realized without the freedom to modify the means by which information is created and acquired. For instance: the number and type of neurons a system possesses, and the manner in which they're connected, which together specify the inputs that can be detected and the actions that can be performed. From this perspective, machine autonomy is about more than the discovery of accurate predictions; it is, among other things, the construction and maintenance of sensors and effectors \cite{Pask1962, Cariani2007, Pharoah2020, Bongard2021}. Our contribution in this paper is a small step toward the realization of such machines.

\section{Materials and Methods.}\label{sec:Section 8}

\subsection{Meta-Learning Algorithm}

We use the model-agnostic Online Meta-Learning algorithm \cite{Javed2019} for learning how to learn over the background partition of the Omniglot image dataset \cite{Lake2015}. This dataset contains 963 hand-written character classes taken from various alphabets. Each character class contains 20 unique instances. All images are resized to 3x28x28 (channels, height, width). A single iteration of meta-learning consists of (\textit{i}) an inner loop of sequential learning over the training images for a single character class $c_{m} \sim \mathcal{C}$; followed by (\textit{ii}) an outer loop that computes accuracy on the inner loop class and a random batch of 64 validation images sampled from the combined set of all other meta-learning classes.

Prediction error on the outer loop following inner loop learning for network $\Theta_{K}^{m}$ is back-propagated through the entire model to update the weights used at the beginning of the inner loop sequence, $\Theta_{0}^{m}$. These new initial weights are then carried over into the next meta-iteration, $\Theta_{0}^{m+1} \propto \Theta_{0}^{m} + \mathrm{update}$. Learning within each inner loop is discarded: only the initial weights are optimized over the outer loop as per \cite{Finn2017}. Successful meta-learners will have learned to learn over the inner loop without corrupting the classification of outer loop images, thus optimizing for the ability to learn without forgetting or interference \cite{Javed2019}. 

Treatments presented in this paper differ both in their architecture, and in which layers are trainable during inner loop meta-learning. Because of this, we make no claims as to the inherent superiority of one model over another. Each treatment was meta-learned for 25,000 meta-iterations on a single NVIDIA Tesla V100 GPU for 25 independent runs having different random seeds. For more details regarding the meta-learning protocol see \cite{Javed2019, Beaulieu2020, Finn2017}.

\subsection{Data Rich/Scarce Meta-Learning}

Data rich meta-learning uses the full background set of Omniglot images (963 character classes) for the OML protocol. The data scarce condition instead uses less than 3$\%$ of the meta-learning data used in the data rich condition (25 randomly sampled character classes). 

\subsection{Domain Transfer}

After executing the meta-learning protocol, each model is trained over a sequence of 100 classes from a previously unseen dataset (ImageNet or CIFAR-100). Classes contain 600 total images, from which 30 are randomly sampled without replacement for each class. This results in 3000 (100*30) sequential iterations of gradient descent. All images are resized to 3x28x28 (channels, height, width). Every run uses a different random seed, a randomized class order, and a random batch of 30 images per class. We applied a grid search over learning rates, and selected the highest performing setting over all subsequent runs.

\subsection{OML}

See \cite{Javed2019, Beaulieu2020} for more details regarding theoretical motivation and network architecture. No changes were made to the meta-learning protocol, except to implement a post-publication correction for computing second-order gradients. The OML treatment, which is distinct from the OML \textit{protocol}, consists of two neural networks: a representation learning network (RLN) composed of five convolutional layers, and a prediction learning network (PLN) composed of two linear layers that takes as input the output of the RLN. During meta-learning, the RLN is updated over the outer loop only, while the PLN is updated in both the inner and outer loops. Under domain transfer, the RLN is fixed over the course of training, while the PLN is fully trainable. Prior to domain transfer, the class prediction layer (CP) of the PLN is randomly reset as per \cite{Javed2019}.

\subsection{TSAR}

TSAR consists of two neural networks: a regulatory network, and a classifier network. The regulatory network consists of (\textit{i}) a perception module; and (\textit{ii}) a regulatory output layer. The regulatory perception module contains 3 convolutional layers, each containing 192 channels (window size=(3,3), stride=1, padding=0). The output of each convolutional layer is followed by instance normalization and a ReLU non-linear activation function. Max-pooling layers (stride=2, kernel size=2) are placed after the first two convolutional layers of the perception module to create an encoding layer of size 1728 from which all regulatory output is generated.

The regulatory output layer consists of four weight matrices, each of which produces a set of regulatory outputs that govern a specific layer in the classification network. A sigmoid activation function is used on regulatory output before modulating synapses in the classifier via multiplicative gating ($\eta^{\ell} \odot W^{\ell}$, for layer $\ell$ in the classifier). 

The classification network is made up of three convolutional layers (112 channels, window size=(3,3), stride=1, padding=0) and a linear output, or class prediction, layer (CP). Max-pooling layers (stride=2, kernel size=2) are placed after each convolutional layer and are followed by instance normalization and a ReLU non-linearity. Class predictions are computed with a softmax function applied to the raw output of the modulated CP layer. During meta-learning, the regulatory network is fixed during inner loop learning, but the prediction network is trainable. All parameters receive updates in the outer loop. 

Under domain transfer, all layers in the prediction network are trainable. The regulatory output layer is also trainable. Only the convolutional layers, or perception module, of the regulatory network are fixed. As with OML, the CP layer, and the regulatory output that flows to this layer, is reset before domain transfer. 

\subsection{Grow and Sculpt}

For \textit{Grow}, all biases in the regulatory output layer are initialized to a value of -8 before executing the meta-learning protocol. Due to the sigmoid activation, this results in initially high levels of synaptic suppression. For \textit{Sculpt}, all biases in the regulatory output layer are initialized to the standard bias of 0 before executing the meta-learning protocol. Thus, all things being equal, synapses in the classification network of Sculpt begin meta-learning with their functional values approximately halved (sigmoid(0.0)=0.5). Under domain transfer, regulatory output to CP is reset with a bias of -2. This value was obtained using a grid search over CP biases and learning rates on domain transfer accuracy for all TSAR models. 

\subsection{Scratch}

This treatment uses the classification network architecture described for TSAR, but with regulation disabled. Scratch is subject to domain transfer starting from a random initialization with no meta-learning.

\subsection{No Regulation}

This treatment uses the classification network architecture described for TSAR, but with regulation disabled. No Regulation is meta-learned on Omniglot using the convolutional layers as the RLN and the class prediction layer as the PLN. Under domain transfer, the PLN is re-initialized and trained over domain transfer sequences, while the RLN is fixed.

\subsection{Reservoir}

This treatment is identical to TSAR (Grow), except that no meta-learning occurs, and regulation is fixed during domain transfer while the classifier is trainable.

\bibliography{pnas-sample}



\break

\begin{figure*}
\centering
\includegraphics[width=1\linewidth]{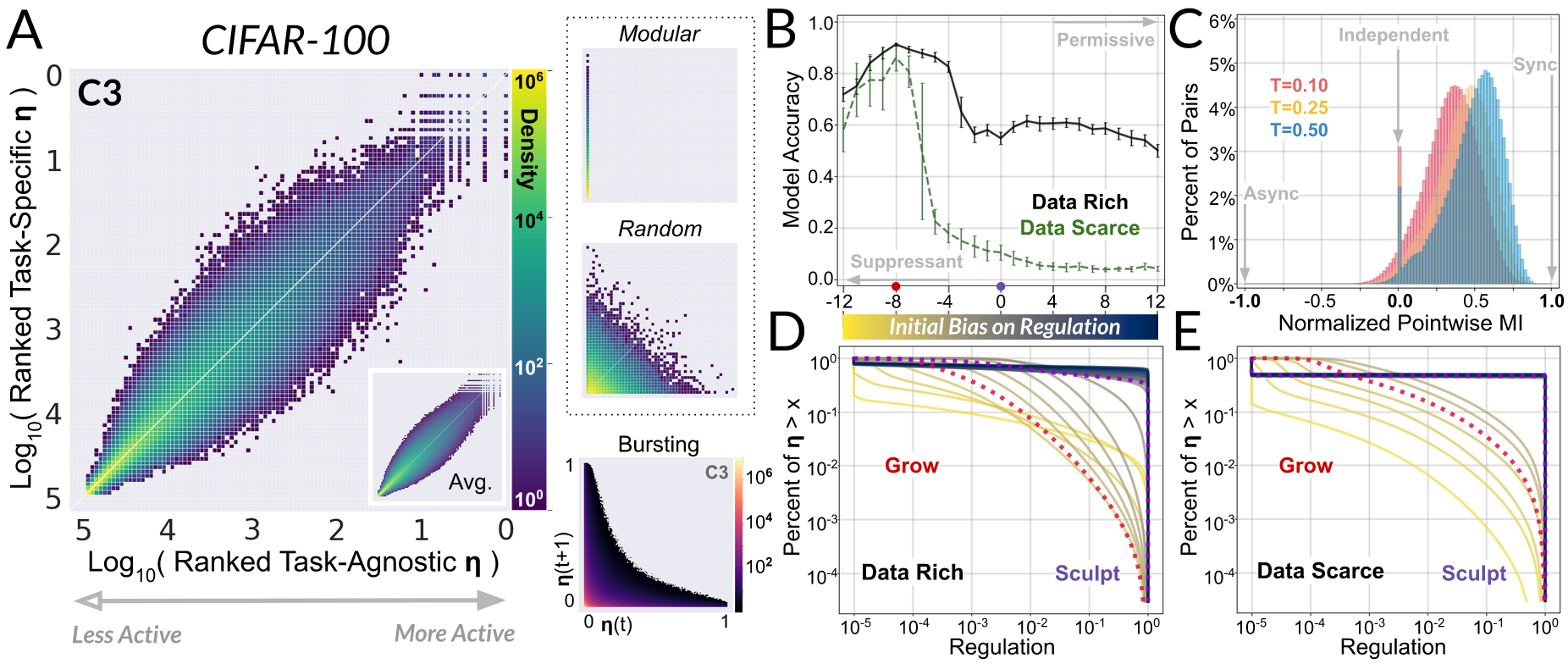}
\caption{Analysis from Figure 3 in the main text for domain transfer to CIFAR-100. Regulation of synapses under domain transfer does not show evidence of task-specific modularity (see hypothetical ``Modular'') but is largely task-agnostic (\textbf{A}). Synapses are likely to be suppressed even if they were highly active on the previous image from the same class (``Bursting''). (\textbf{B}) Performance under domain transfer to CIFAR-100 for the range of initial biases considered. Standard deviation is reported for error bars. (\textbf{C}) Normalized pointwise mutual information \cite{churchhanks} for each pair of synapses receiving regulation above the given threshold for more than 1\% of images. We find a positive bias in the degree of correlation among synapse pairs, but an absence of strong synchronicity or asynchronicity. This indicates that the coalition of weights used for a given image is relatively heterogeneous with respect to those used for other images. Increasingly scale-free cumulative distribution functions for regulation obtain under domain transfer for increasingly successful regulators after data-scarce (\textbf{E}) and, especially, data-rich (\textbf{D}) meta-learning.  }
\label{fig:appendix1}
\end{figure*}

\begin{figure*}
\centering
\includegraphics[width=1\linewidth]{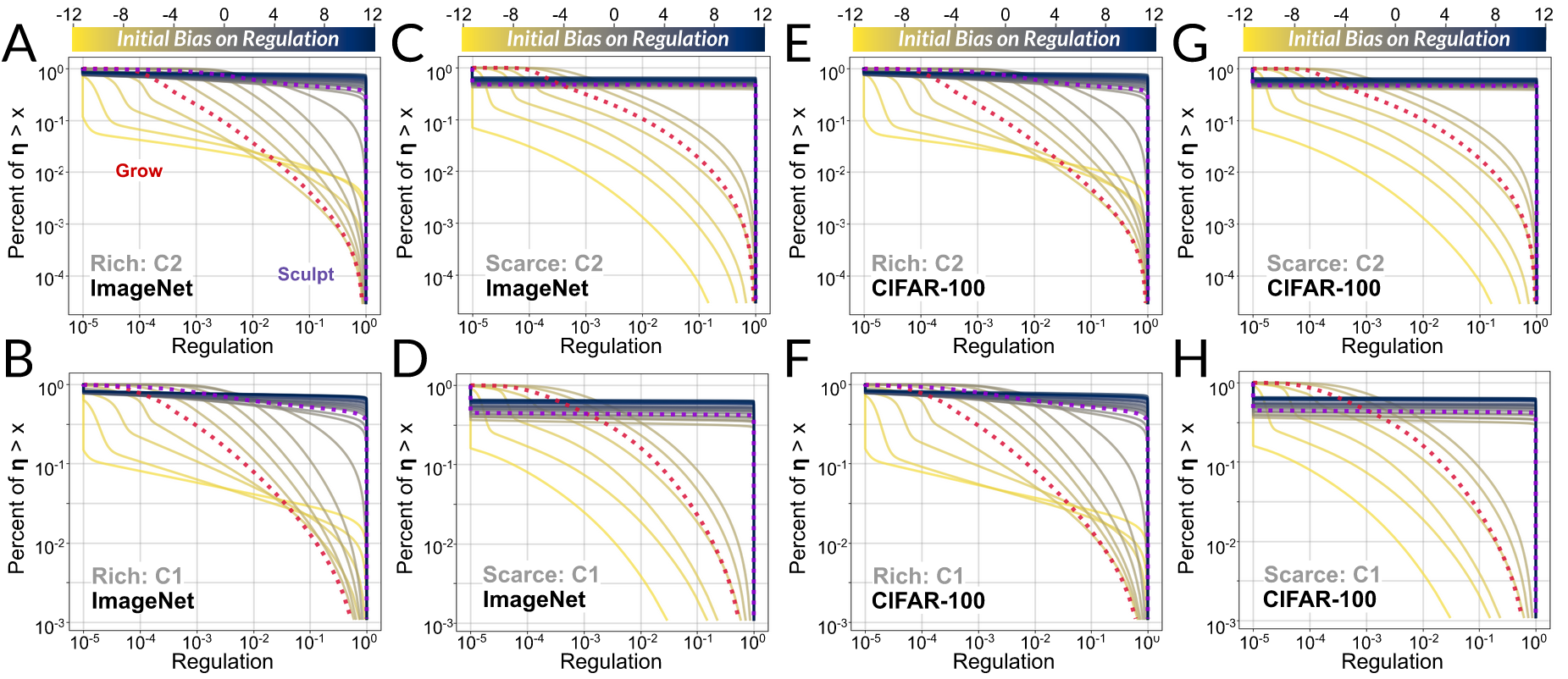}
\caption{\textbf{A-H} For data rich and data scarce meta-learning, regulators that are increasingly successful under domain transfer exhibit regulatory distributions that are increasingly linear (scale-free) in the logarithmic CCDF \cite{Stumpf2012}. This trend holds across several orders of magnitude in all convolutional layers of the Grow condition. However, our results do not hinge on whether regulation is precisely power-law distributed, or merely log-normally distributed. These results shows that Grow has obtained a balance of activity and suppression that is analogous to the balance of order and disorder in critical systems at the edge of chaos \cite{Kauffman1991}. We present CCDF plots for the highest performing runs for each model.}
\label{fig:appendix2}
\end{figure*}

\begin{figure*}
\centering
\includegraphics[width=1\linewidth]{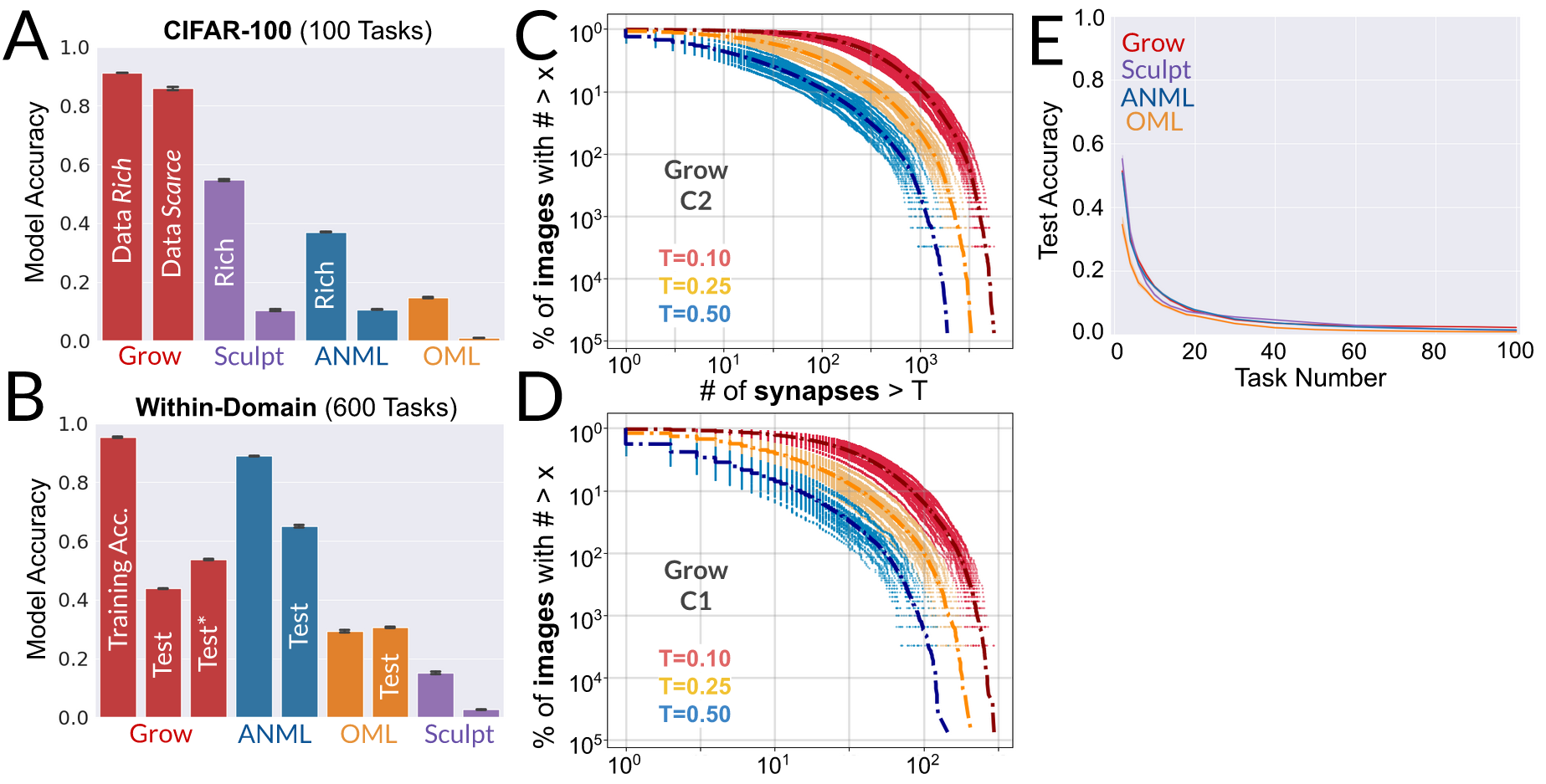}
\caption{(\textbf{A}) Training accuracy under domain transfer to CIFAR-100. \textbf{Rich}: Grow=91.1\% $\pm{0.56}$\% (SD), Sculpt=54.7\% $\pm{2.4}$\%, ANML=37\% $\pm{1.70}$\%, OML=14.6\% $\pm{2.6}$\%. \textbf{Scarce}: Grow=85.9\% $\pm{5.12}$\%, Sculpt=10.4\% $\pm{2.9}$\%, ANML=10.57\% $\pm{0.73}$\%, OML=1\% $\pm{0.10}$\%. Accuracy for 600 previously unseen tasks learned sequentially from Omniglot. \textbf{Training}: Grow=95.2\% $\pm{1.4}$\%, Sculpt=15.1\% $\pm{2.23}$\%, ANML=88.9\% $\pm{0.59}$\%, OML=29.3\% $\pm{2.23}$\%. \textbf{Validation}: Grow=43.8.2\% $\pm{0.93}$\%, Sculpt=2.6\% $\pm{0.28}$\%, ANML=65.06\% $\pm{2.88}$\%, OML=30.6\% $\pm{1.04}$\%. (\textbf{B}) Test accuracy on held-out Omniglot images (600 classes; 15 training images/5 test images per class). Although our method under-performs ANML, only the final layer of the classification network in ANML is trainable. ANML also uses a different network architecture than TSAR, and so is not directly comparable. When the regulatory output layers that control C1, C2, and C3 in Grow are fixed, but the weights they regulate are trainable, test accuracy ("Grow", 3rd bar) increases to 53.75\% $\pm{1.45}$\%. This final version of Grow corresponds to a relative drop of 15.5\% compared to the IID setting (epochs=3: 63.5\%). (\textbf{C}, \textbf{D}) Complementary cumulative distribution function (CCDF) for the percent of images in ImageNet that cause a given number of synapses to receive regulation above a threshold (T). We report individual runs (dot) and the mean across runs (dash-dot). (\textbf{E}) Test accuracy on ImageNet. These results indicate that the problem of forgetting may be orthogonal to the problem of generalization. (\textbf{F}) The regulator toggles between two states, each with a similar presentation across tasks (\textit{enhancing} and \textit{diminishing}). Points in the right panel are dimensionally reduced regulator encodings for the images that are the most enhancing (top 256, yellow-red) and most diminishing (bottom 256, grey-black). Coloring by task identity (left panel) yields no discernable clusters, and for no run are tasks identifiable by their t-SNE projections using K-Nearest Neighbors. Dimensional reduction is obtained through a combination of principle component analysis (50 components) and the t-SNE algorithm \cite{vandermaaten08}.}
\label{fig:appendix3}
\end{figure*}

\begin{figure*}
\centering
\includegraphics[width=1\linewidth]{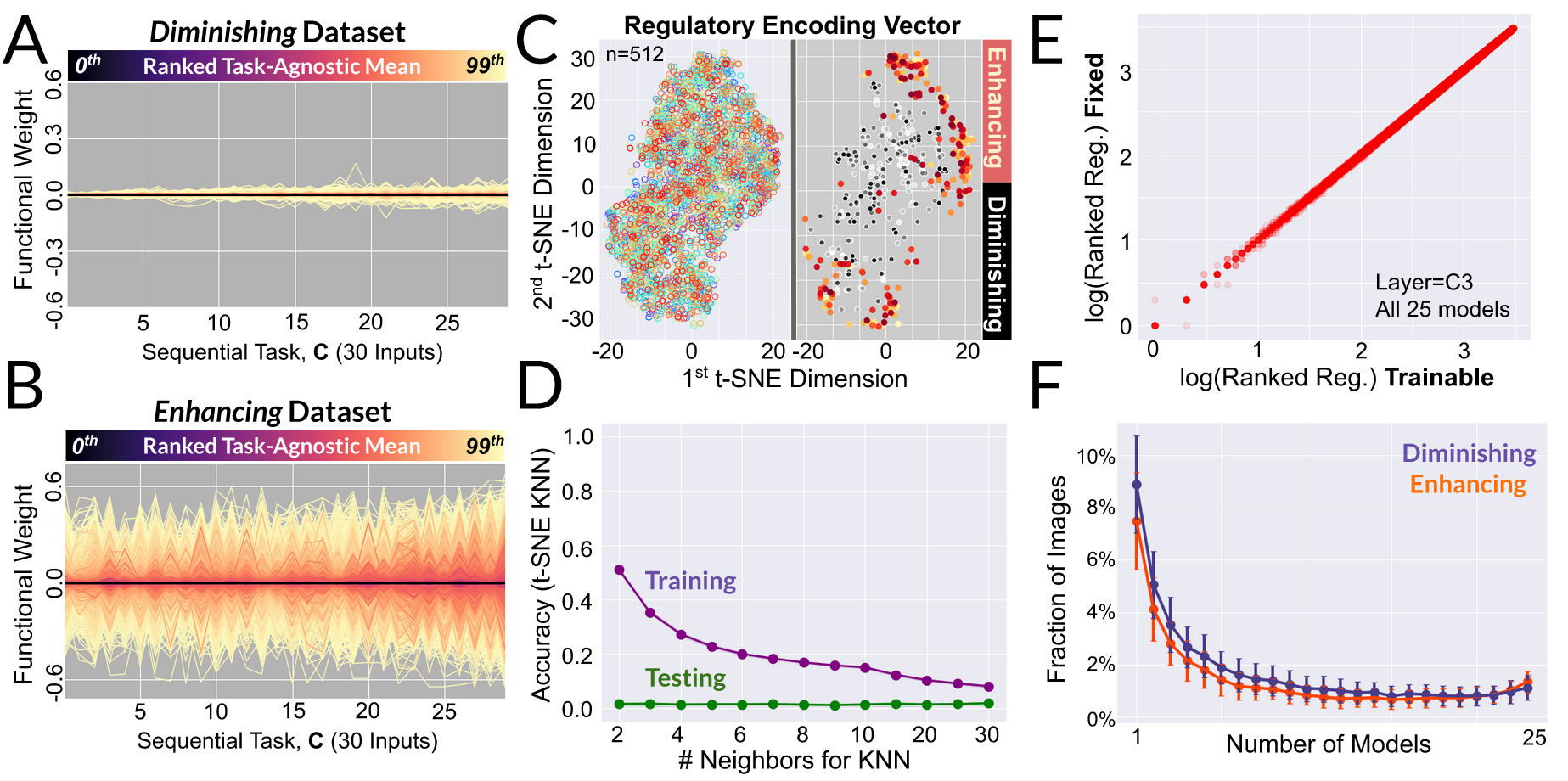}
\caption{Randomly sampled window of synaptic activity (post-masking) under domain transfer for Grow in the Diminishing (\textbf{A}) and Enhancing (\textbf{B}) datasets. (\textbf{C}) Dimensionally C3 activations are not identifiable by their class ID. Following dimensional reduction (D=2) through a combination of PCA (50 components) and t-SNE \cite{vandermaaten08}, K-Nearest Neighbors for class prediction was performed. These results show that images belonging to a particular class are not represented by the classifier in a way that is more similar than images belonging to other classes. Confidence intervals report standard deviation across all runs (n=25) (\textbf{D}) Dimensionally reduced regulatory encoding vectors are not identifiable by their class ID. Following dimensional reduction (D=2) through a combination of PCA (50 components) and t-SNE \cite{vandermaaten08}, K-Nearest Neighbors for class prediction was performed. These results show that images belonging to a particular class are not represented by the regulator in a way that is more similar than images belonging to other classes. Confidence intervals report standard deviation across all runs (n=25). (\textbf{E}) Log-ranked regulation for each image in ImageNet across 25 models of Grow when regulation is trainable (x-axis) and when it is fixed (y-axis). (\textbf{F}) For each class in Imagenet, all 600 images are ranked according to mean regulatory output for all 25 models of Grow. We report the mean fraction of images with membership in the top/bottom 100 images over all classes for the corresponding number of models (x-axis). Thus, the leftmost result reads that an average of 9\% of images belong to the diminishing set (bottom 100) of just 1 model. Enhancing and diminishing sets used in the main text consider only the top 30 images per class. Error bars report standard deviation. (\textbf{C}).}
\label{fig:appendix4}
\end{figure*}

\begin{figure*}
\centering
\includegraphics[width=1\linewidth]{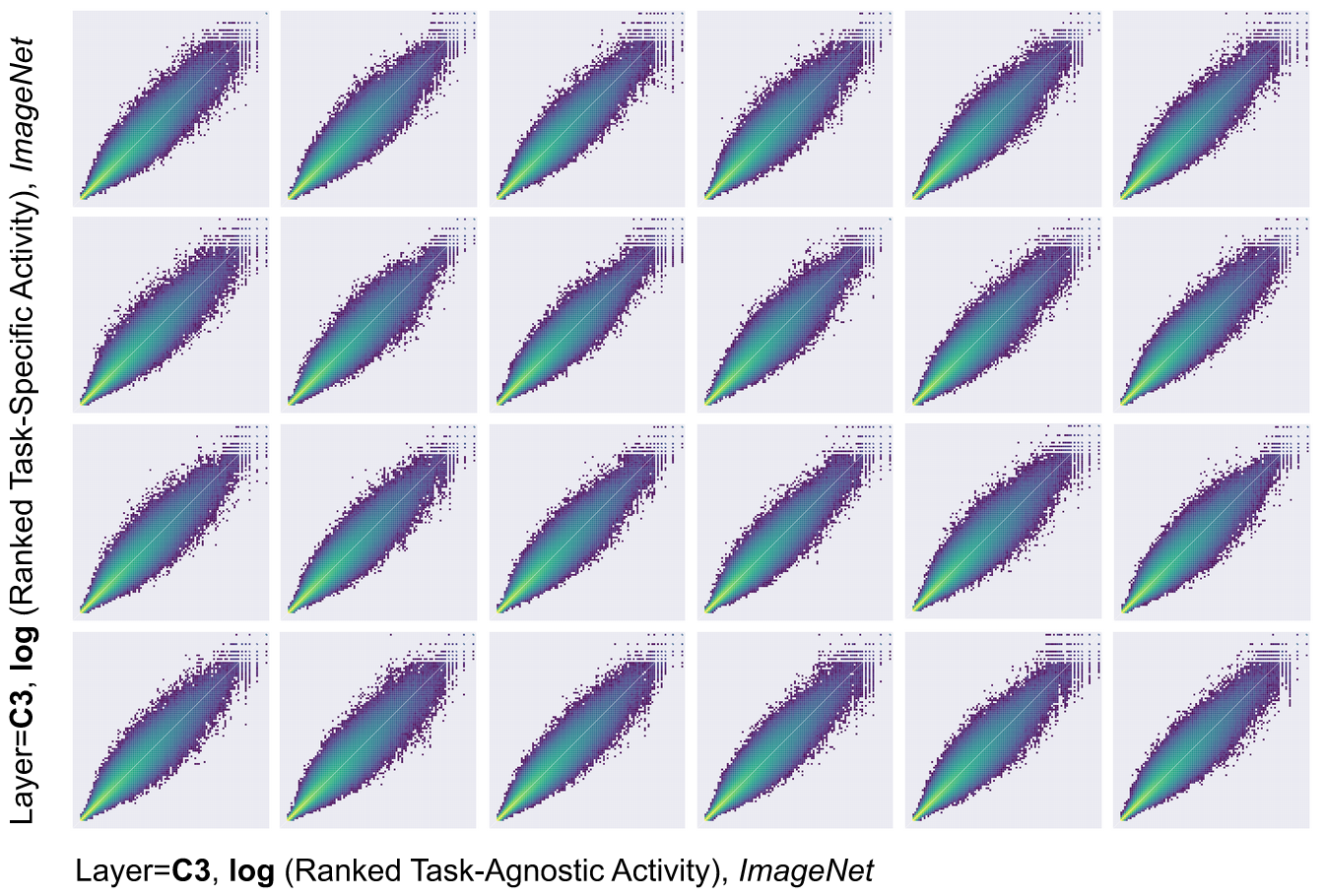}
\caption{Regulation in C3 of the data rich condition under domain transfer to ImageNet does not elicit task-specific modularity. Instead, the most active weights on a given task are the most active weights over all tasks (\textit{synaptic recycling}), and no synapse dramatically changes rank for any individual task. Here we present results for a single trial for each of the remaining 24 models. Finally, we note that the degree of context \textit{sensitivity} is higher for layer C3 under domain transfer to ImageNet and CIFAR-100 than it is for layers C2 and C1. This is distinct from context \textit{modularity}, and likely reflects the commonly observed property that downstream layers are dedicated to less generic features of the input.}
\label{fig:appendix5}
\end{figure*}

\begin{figure*}
\centering
\includegraphics[width=1\linewidth]{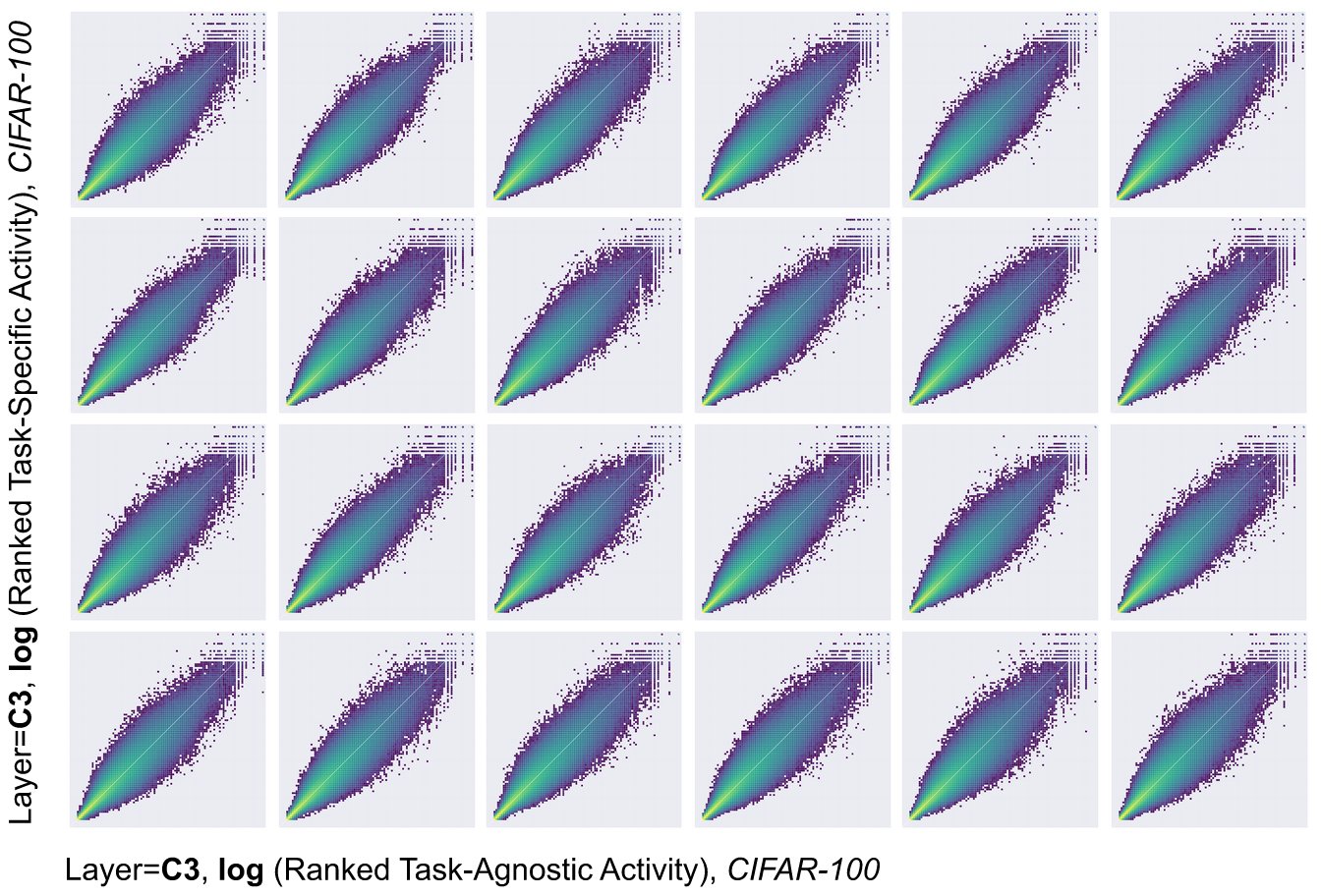}
\caption{Regulation in C3 of the data rich condition under domain transfer to CIFAR-100 does not elicit task-specific modularity. Instead, the most active weights on a given task are the most active weights over all tasks (\textit{synaptic recycling}), and no synapse dramatically changes rank for any individual task. Here we present results for a single trial for the remaining 24 independent models. Finally, we note that the degree of context \textit{sensitivity} is higher for layer C3 under domain transfer to ImageNet and CIFAR-100 than it is for layers C2 and C1. This is distinct from context \textit{modularity}, and likely reflects the commonly observed property that downstream layers are dedicated to less generic features of the input.}
\label{fig:appendix6}
\end{figure*}

\begin{figure*}
\centering
\includegraphics[width=1\linewidth]{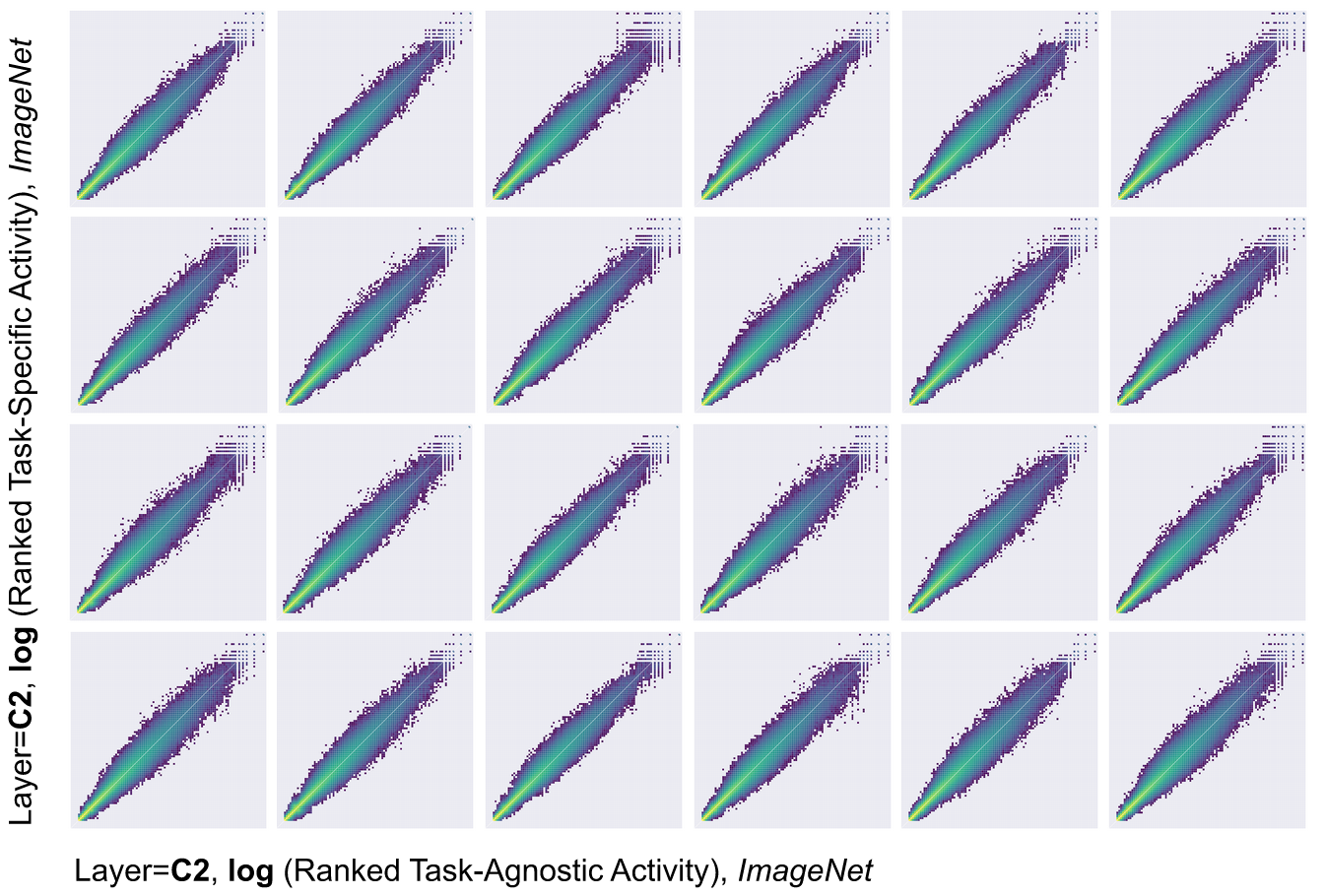}
\caption{Regulation in C2 of the data rich condition under domain transfer to ImageNet does not elicit task-specific modularity. Instead, the most active weights on a given task are the most active weights over all tasks (\textit{synaptic recycling}), and no synapse dramatically changes rank for any individual task. Here we present results for a single trial for 24 independent models. Compared to results for regulation of C3 under domain transfer to ImageNet, regulation of C2 is noticeably less context-sensitive. This may be attributed to the common observation that successive layers in neural networks attend to decreasingly generic properties of inputs. Nevertheless, context-dependent modules, which are an extreme form of context sensitivity, are not present in any of the convolutional layers.}
\label{fig:appendix7}
\end{figure*}

\begin{figure*}
\centering
\includegraphics[width=1\linewidth]{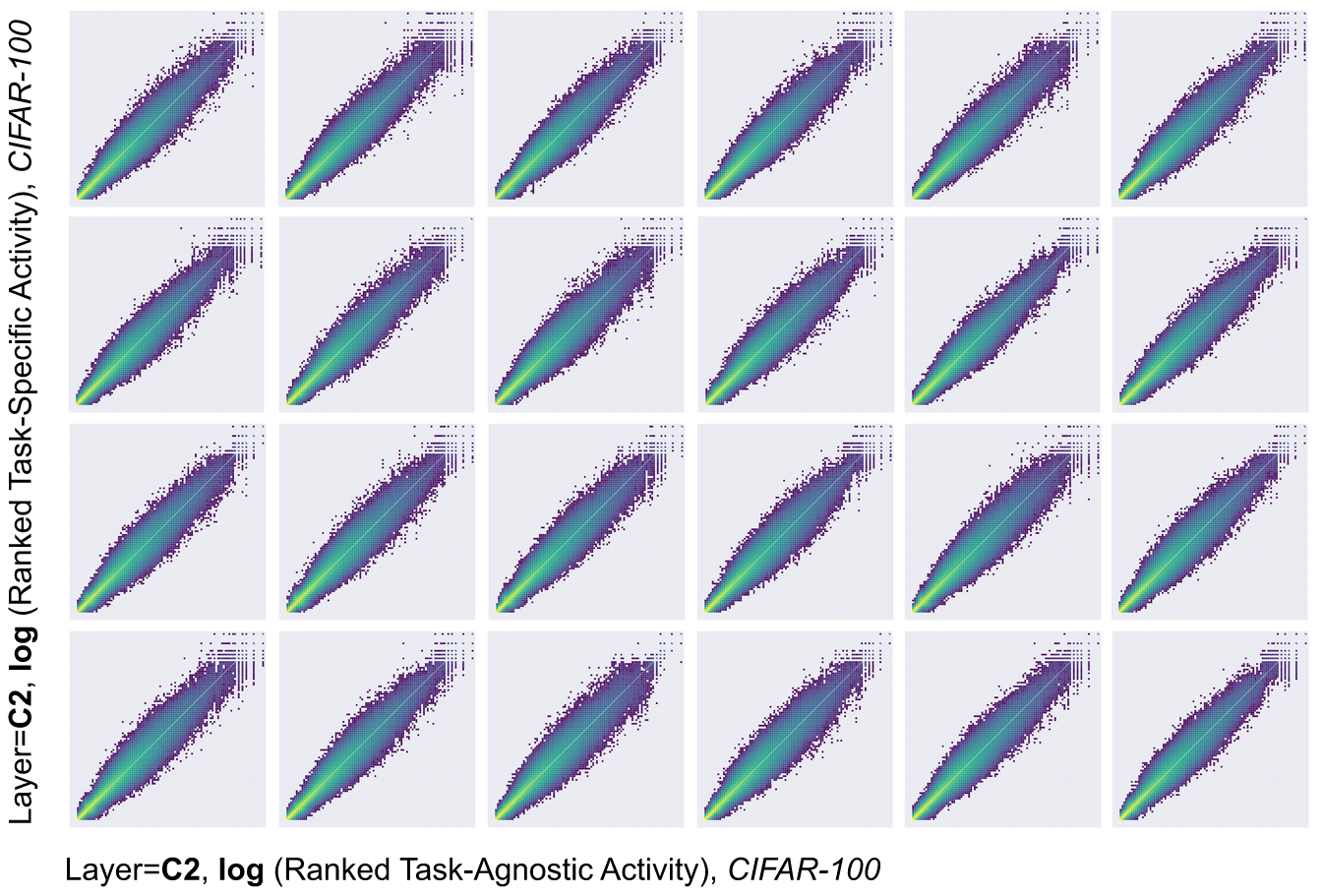}
\caption{Regulation in C2 of the data rich condition under domain transfer to CIFAR-100 does not elicit task-specific modularity. Instead, the most active weights on a given task are the most active weights over all tasks (\textit{synaptic recycling}), and no synapse dramatically changes rank for any individual task. Here we present results for a single trial for each of the remaining 24 models. Compared to results for regulation of C3 under domain transfer to CIFAR-100, regulation of C2 is noticeably less context-sensitive. This may be attributed to the common observation that successive layers in neural networks attend to decreasingly generic properties of inputs. Nevertheless, context-dependent modules, which are an extreme form of context sensitivity, are not present in any of the convolutional layers.}
\label{fig:appendix8}
\end{figure*}

\begin{figure*}
\centering
\includegraphics[width=1\linewidth]{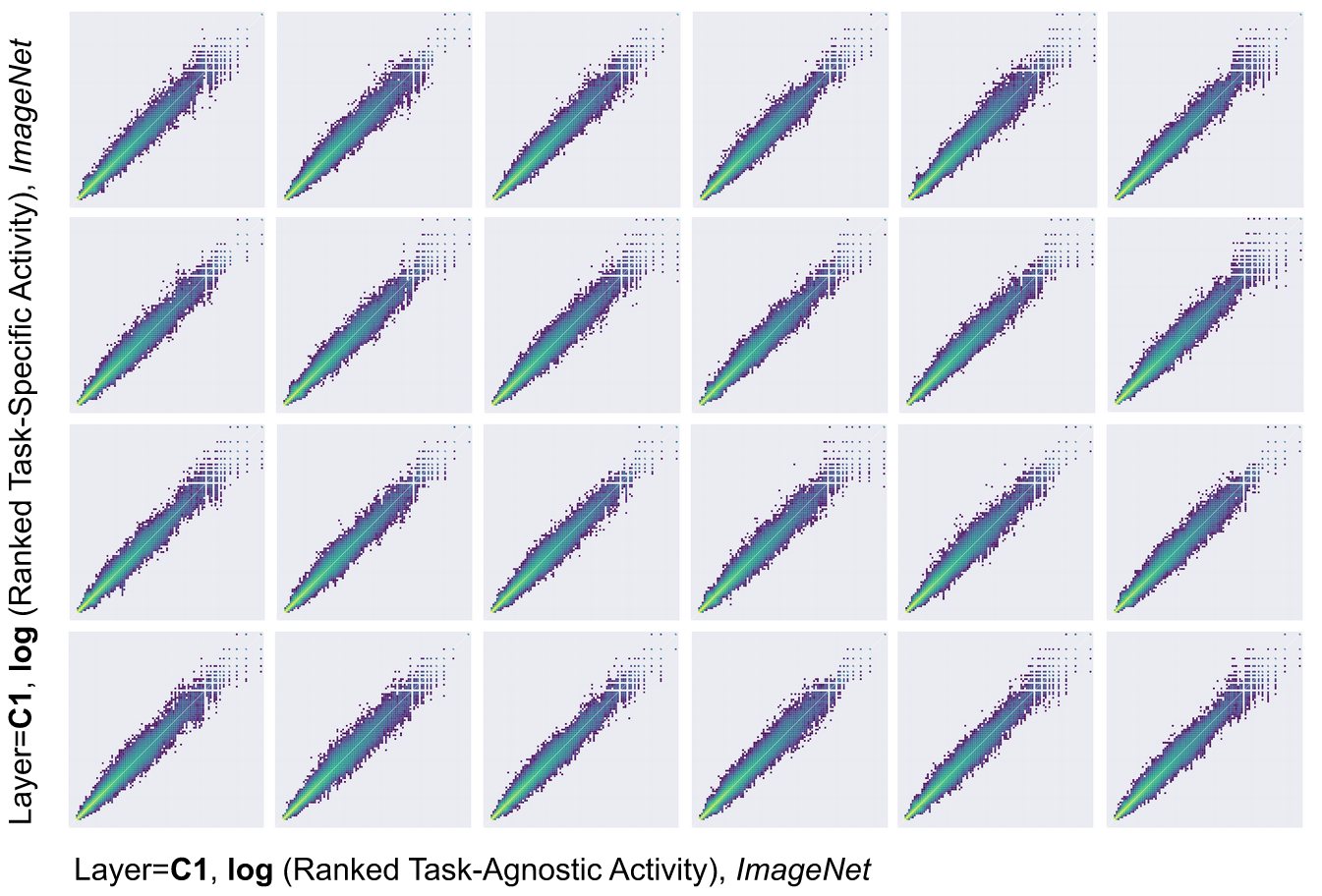}
\caption{Regulation in C1 of the data rich condition under domain transfer to ImageNet does not elicit task-specific modularity. Instead, the most active weights on a given task are the most active weights over all tasks (\textit{synaptic recycling}), and no synapse dramatically changes rank for any individual task. Here we present results for a single trial for each of the remaining 24 models. Compared to results for regulation of C3 and C2 under domain transfer to ImageNet, regulation of C1 is noticeably less context-sensitive. This may be attributed to the common observation that successive layers in neural networks attend to decreasingly generic properties of inputs. Nevertheless, context-dependent modules, which are an extreme form of context sensitivity, are not present in any of the convolutional layers of the classifier. }
\label{fig:appendix9}
\end{figure*}

\begin{figure*}
\centering
\includegraphics[width=1\linewidth]{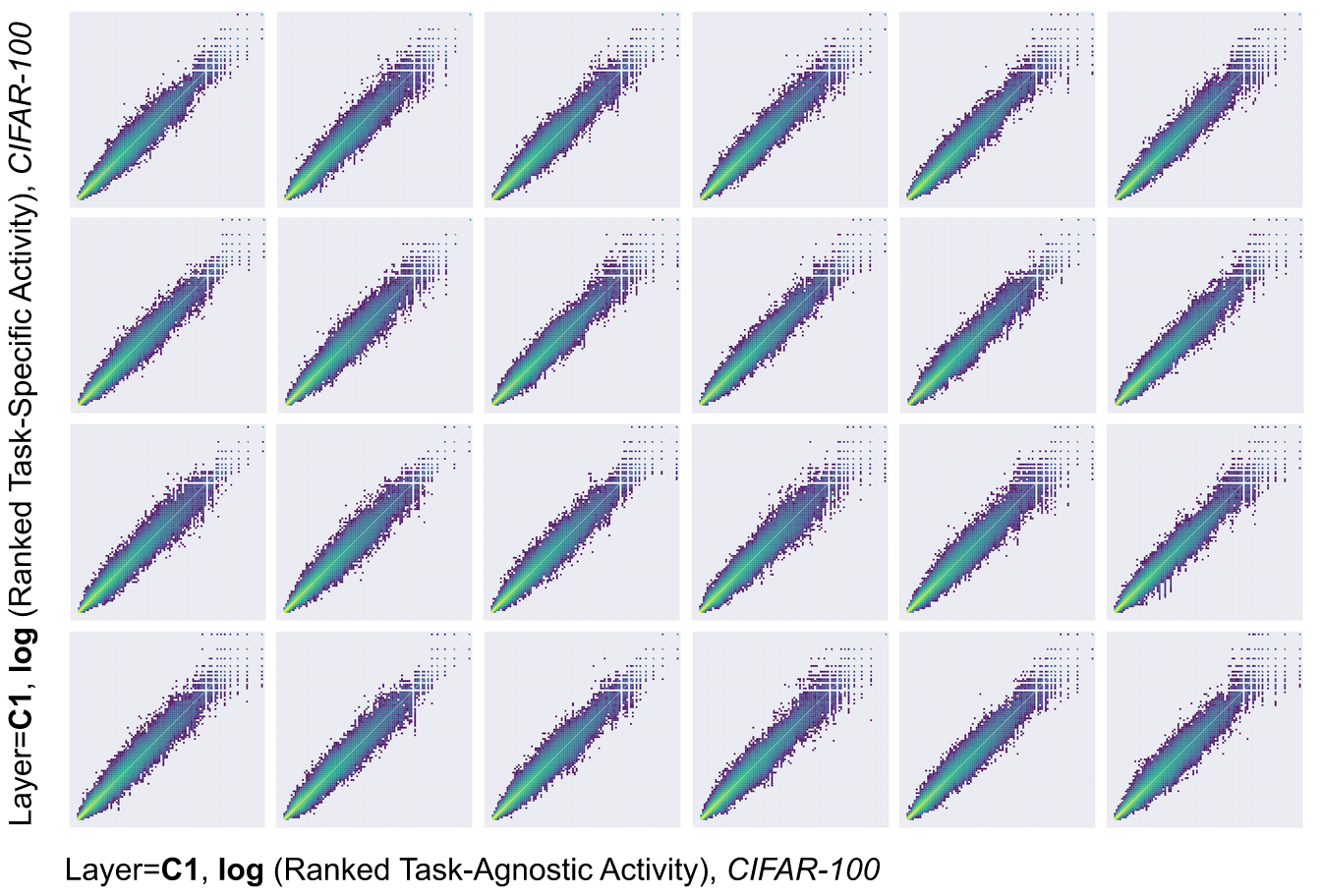}
\caption{Regulation in C1 of the data rich condition under domain transfer to CIFAR-100 does not elicit task-specific modularity. Instead, the most active weights on a given task are the most active weights over all tasks (\textit{synaptic recycling}), and no synapse dramatically changes rank for any individual task. Here we present results for a single trial for each of the remaining 24 models. Compared to results for regulation of C3 and C2 under domain transfer to CIFAR-100, regulation of C1 is noticeably less context-sensitive. This may be attributed to the common observation that successive layers in neural networks attend to decreasingly generic properties of inputs. Nevertheless, context-dependent modules, which are an extreme form of context sensitivity, are not present in any of the convolutional layers of the classifier.}
\label{fig:appendix10}
\end{figure*}

\begin{figure*}
\centering
\includegraphics[width=1\linewidth]{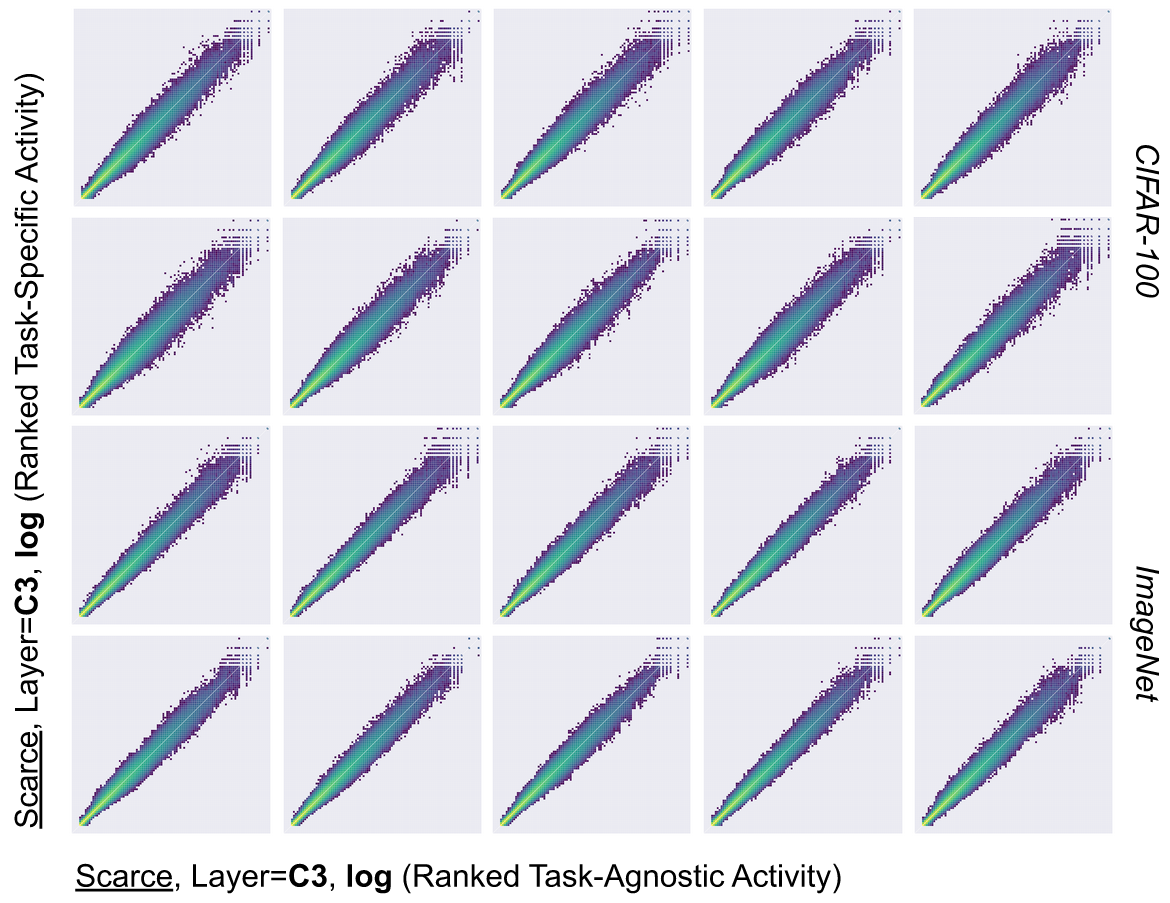}
\caption{The lack of task-specific modularity observed in the regulation of C3 in the data rich condition is recapitulated in regulation of C3 in the data scarce condition. We also find that regulation in the data scarce condition is noticeably less context sensitive than in the data rich condition. Thus, we find that synaptic recycling in C3 is more strongly pronounced in the data scarce condition. We present results for 10 randomly sampled independent runs per dataset.}
\label{fig:appendix11}
\end{figure*}

\begin{figure*}
\centering
\includegraphics[width=1\linewidth]{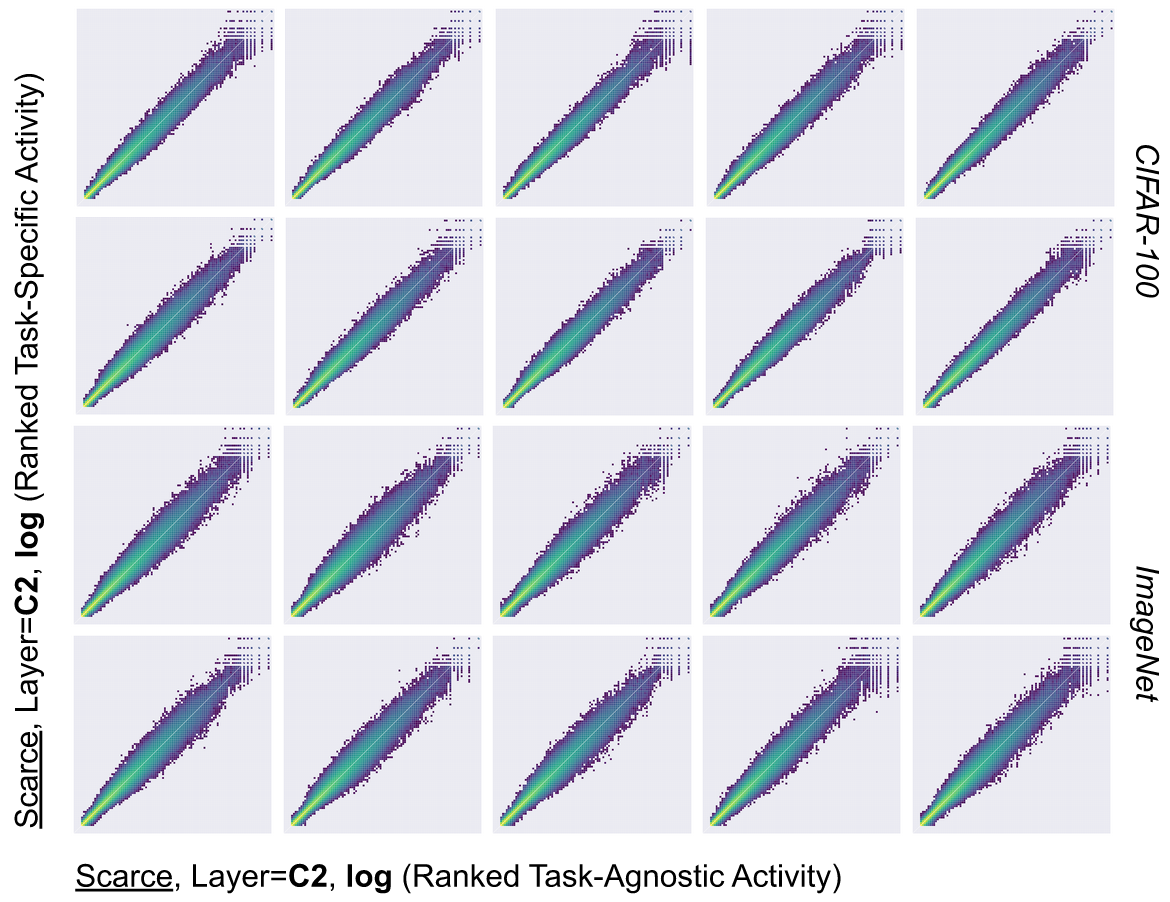}
\caption{The lack of task-specific modularity observed in the regulation of C2 in the data rich condition is recapitulated in regulation of C2 in the data scarce condition. We also find that regulation in the data scarce condition is noticeably less context sensitive than in the data rich condition. Thus, we find that synaptic recycling in C2 is more strongly pronounced in the data scarce condition. We present results for 10 randomly sampled independent runs per dataset.}
\label{fig:appendix12}
\end{figure*}

\begin{figure*}
\centering
\includegraphics[width=1\linewidth]{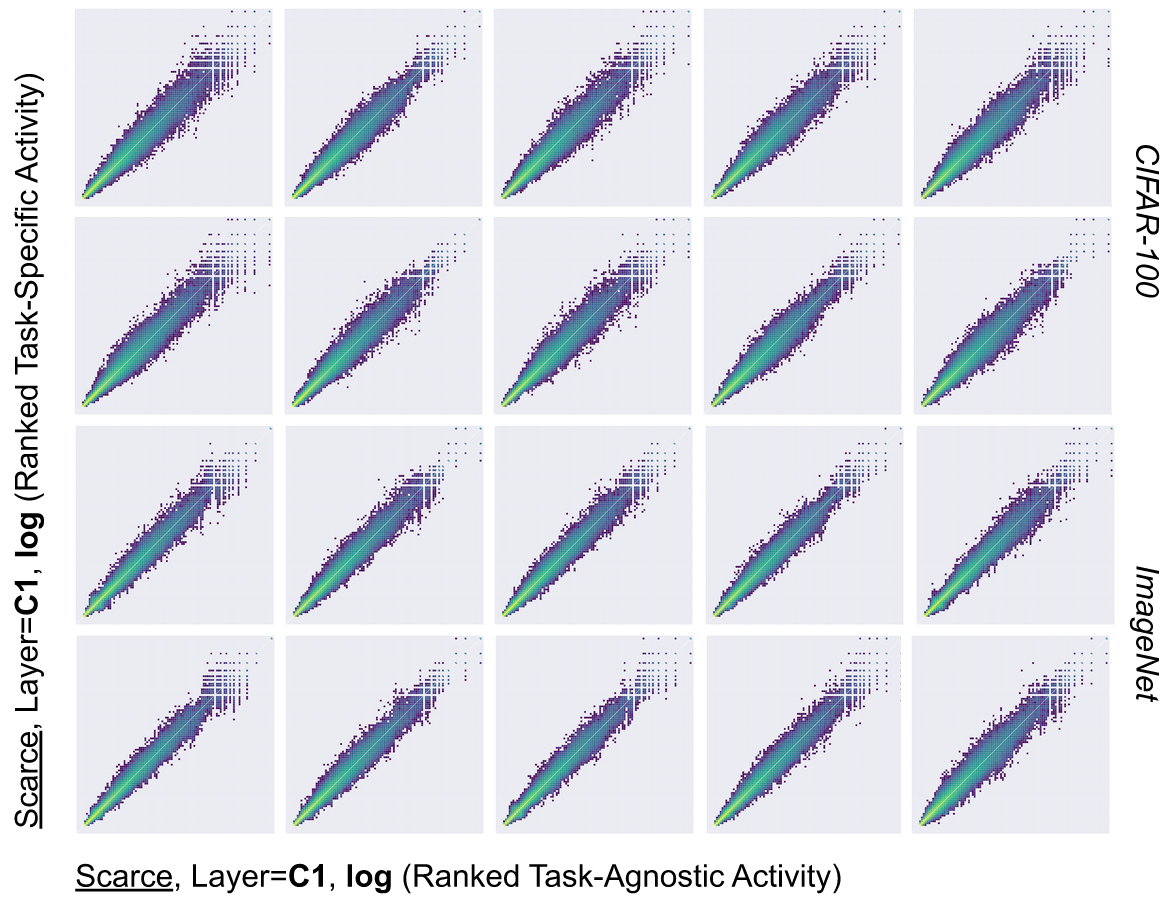}
\caption{The lack of task-specific modularity observed in the regulation of C1 in the data rich condition is recapitulated in regulation of C1 in the data scarce condition. We also find that regulation in the data scarce condition is noticeably less context sensitive than in the data rich condition. Thus, we find that synaptic recycling in C1 is more strongly pronounced in the data scarce condition. We present results for 10 randomly sampled independent runs per dataset.}
\label{fig:appendix13}
\end{figure*}

\end{document}